Thesis for the Degree of Doctor

# Interpretable temporal fusion network of multi- and multi-class arrhythmia classification

by

Yun Kwan Kim

Department of
Brain and Cognitive Engineering

Korea University

February 2026

博 士 學 位 論 文

# Interpretable temporal fusion network of multi- and multi-class arrhythmia classification

高麗大學校 大學院
腦工學科
金 倫 寬

2026 年 2 月

李 晟 瑍   敎授指導
博 士 學 位 論 文

# Interpretable temporal fusion network of multi- and multi-class arrhythmia classification

이 論文을 工學 博士學位 論文으로 提出함

2026 年 2月

高麗大學校 大學院
腦工學科
金 倫 寬

金 倫 寬의 工學 博士學位 論文 審査를 完了함

2026 年 2 月

委員長＿＿＿이 성 환＿＿＿（印）

委　員＿＿＿정 원 주＿＿＿（印）

委　員＿＿＿감 태 의＿＿＿（印）

委　員＿＿＿정 지 채＿＿＿（印）

委　員＿＿＿설 상 훈＿＿＿（印）

# Abstract

Clinical decision support systems (CDSSs) have been widely utilized to support the decisions made by cardiologists when detecting and classifying arrhythmia from electrocardiograms. However, forming a CDSS for the arrhythmia classification task is challenging due to the varying lengths of arrhythmias. Although the onset time of arrhythmia varies, previously developed methods have not considered such conditions. Thus, we propose a framework that consists of (i) local and global extraction and (ii) local-global information fusion with attention to enable arrhythmia detection and classification within a constrained input length. The framework's performance was evaluated in terms of 10-class and 4-class arrhythmia detection, focusing on identifying the onset and ending point of arrhythmia episodes and their duration using the MIT-BIH arrhythmia database (MITDB) and the MIT-BIH atrial fibrillation database (AFDB). Duration, episode, and Dice score performances resulted in overall F1-scores of 96.45%, 82.05%, and 96.31% on the MITDB and 97.57%, 98.31%, and 97.45% on the AFDB, respectively. The results demonstrated statistically superior performance compared to those of the benchmark models. To assess the generalization capability of the proposed method, an MITDB-trained model and MIT-BIH malignant ventricular arrhythmia database-trained model were tested AFDB and MITDB, respectively. Superior performance was attained compared with that of a state-of-the-art model. The proposed method effectively captures both local and global information and dynamics without significant information loss. Consequently, arrhythmias can be detected with greater accuracy, and their occurrence times can be precisely determined, enabling the clinical field to develop more accurate treatment plans based on the proposed method.

# 요약

임상의사결정지원시스템(CDSS)은 심전도(ECG)를 기반으로 부정맥을 탐지하고 분류하는 과정에서 심장 전문의의 의사결정을 지원하는 데 폭넓게 활용되어 왔다. 그러나 부정맥 분류 과제는 발생 기간의 다양성 및 발현 시점 차이 등으로 인해 CDSS 구축에 어려움이 있다. 기존의 방법론들은 부정맥의 시작 시점 변이를 충분히 고려하지 못했다. 이에 본 연구는 (i) 국소 및 전역 정보 추출, (ii) 주의(attention) 기반 국소-전역 정보 융합 두 가지 전략을 결합한 새로운 프레임워크를 제안한다. 제안한 프레임워크는 MIT-BIH 부정맥 데이터베이스(MITDB)와 MIT-BIH 심방세동 데이터베이스(AFDB)를 이용하여 10개 및 4개 클래스에 대해 부정맥 에피소드의 시작·종료 시점과 전체 기간을 정확하게 탐지 및 분류하는지 성능을 평가하였다.

MITDB에 대한 전반적 정확도(F1-score)는 기간·에피소드·Dice 지표 기준으로 각각 96.45%, 82.05%, 96.31%였고, AFDB에서는 97.57%, 98.31%, 97.45%를 기록하였다. 모든 결과는 벤치마크 모델 대비 통계적으로 유의하게 우수한 성능을 보였다. 모델의 범용성 평가를 위해 MITDB, MADB, AFDB 간 교차 테스트를 수행한 결과, 최신 비교 모델 대비 뛰어난 성능을 입증하였다. 제안한 방법은 국소 및 전역 정보와 시계열의 동적 특성을 효과적으로 포착하여 정보 손실 없이 부정맥을 탐지할 수 있다. 결과적으로, 제안 모델은 부정맥의 정확한 진단과 발생 시점 산출을 가능케 하며, 임상 현장에서 환자 맞춤 치료계획 수립에 기여할 수 있다.

# Contents

# List of Figures

# List of Tables

# 1  Introduction

Clinical decision support systems (CDSSs) are medical information technology systems designed to provide clinical decision support using a knowledge base; such systems apply rules to patient data and employ machine learning to analyze clinical data [1]. CDSSs have been developed to utilize electrocardiograms (ECGs), capitalizing on various applications, including Holter monitoring systems, real-time patient monitoring, and heart failure prediction [2]. Holter monitoring systems, which require comprehensive analysis of extensive ECG recordings, have incorporated arrhythmia detection and classification algorithms [2]. Arrhythmia is a progressive condition where brief episodes can often evolve from intermittent to continuous states of arrhythmia [3]. Detecting, classifying, and evaluating the patterns of these episodes is crucial, as they may exhibit instability and require different treatment strategies [4]. These computer-aided arrhythmia classification algorithms present new possibilities for cardiologists to achieve improved diagnostic precision [5].

The ECG records the heart's electrical activity through electrodes placed on the skin. ECGs are extensively used in clinical settings for arrhythmia detection. Arrhythmias are characterized by specific features, including their sudden onset and termination [6]. Additionally, arrhythmias can persist over extended periods [7]. Cardiologists monitor changes in ECG rhythms to assess cardiovascular pathology and provide critical reference information for cardiac diagnosis [8]. Continuous Holter recordings over 24 hours are essential within CDSSs to accurately detect arrhythmias. However, the manual analysis and interpretation of ECGs is a time-intensive and laborious task, particularly when ECGs span long durations. To mitigate these challenges, various studies have focused on developing CDSSs that use automated algorithms to enhance arrhythmia detection and improve diagnostic accuracy [9, 10].

Holter monitoring is capable of measuring the burden of arrhythmia, defined as the percentage of time an arrhythmia occurs within a specified period. This measurement allows medical practitioners to evaluate the effectiveness of diagnostic and rhythm control strategies, making precise burden calculation essential. To determine this burden, it is critical to obtain information about the duration of each arrhythmia, including its onset and ending points. For an automated system to analyze arrhythmia burden, the initial occurrence point of the ground truth arrhythmia needed to be identified in a 10-second segmented dataset. This onset point was then paired with the final point of the arrhythmia that corresponded to this onset, which was designated as the ending point [11, 12]. The duration and episodes of arrhythmia within each 10-second ECG segment were then calculated using the defined onset and ending points.

Many studies have utilized machine learning methods using preprocessing, feature extraction, and feature segmentation techniques for automated arrhythmia classification [13]. To build robust Holter monitoring systems for arrhythmia classification, many researchers have adopted deep learning approaches [9, 11, – 13, 14], resulting in significant performance improvements. However, many studies, including the ones mentioned above, have approached classification with the assumption that only one arrhythmia is present in the input data.

Teplitzky et al. [15] proposed a ResNet-based sequence-to-sequence architecture for arrhythmia classification across 14-class, 11-class, 7-class, and 5-class configurations, achieving F1-scores of 70%, 80%, 90%, and 95%, respectively. This approach allowed for the classification of arrhythmias of varying durations using sequence-level inference with 1-second moving intervals. However, their method primarily focused on local arrhythmic patterns and did not account for global patterns in the input, such as paroxysmal arrhythmia, ventricular tachycardia (VT), and atrial fibrillation (AFIB) [16]. Consequently, there is a need for accurate automated arrhythmia detection systems capable of considering both local and global patterns to improve diagnostic utility in clinical settings.

Computer-aided CDSSs, including Holter monitoring and real-time arrhythmia monitoring systems in the real world, set a predefined input window and analyze the output in the corresponding input window. Additionally, the system described in [16, 18] employed a sliding window approach, where results were generated as the predefined input window shifted incrementally [19]. These systems typically perform arrhythmia classification under the assumption that each input segment yields a single output. However, due to the complex nature of cardiac dynamics, a specific segmented input signal may contain multiple events, as illustrated in Figure 1.

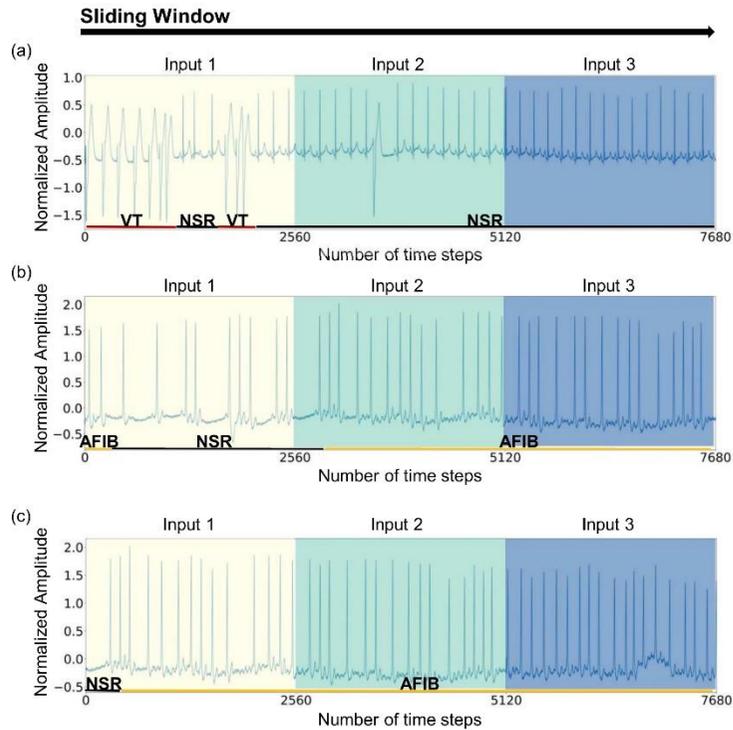

Figure 1. Representative examples of arrhythmia characteristics from the MITDB. Each example shows the ECG signals of arrhythmia characteristics, including abrupt changes and persistent progression, when using the sliding window method. Each input window is 10-second long, and the color of each input is expressed differently. Next, at the bottom, the red horizontal bar indicates VT, the black horizontal bar represents NSR, and finally, the yellow horizontal bar shows AFIB. (a) VT and NSR. (b) AFIB and NSR. (c) NSR and AFIB.

While deep learning-based approaches have demonstrated promising results in arrhythmia classification, challenges persist for AI-based applications in Holter monitoring, real-time patient monitoring, and smart wearable devices. Although previous studies have proposed methods to enhance multiclass classification for a single output, detecting multiple arrhythmias within a predetermined window remains a significant challenge. Multiple arrhythmias can present as various distinct ECG patterns within a single input window, as illustrated in Figure 1, complicating the classification process. For example, in an input ECG signal of 10-second, VT may appear within 0 to 4-second, normal sinus rhythm (NSR) may appear within 4 to 6-second, VT may reappear within 6 to 8-second, and NSR may reappear within 8 to 10-second, as shown in Fig. 1(a). There are four events, including VT and NSR, indicating that multiple arrhythmia events can occur within a 10-second input segment. Additionally, "multiclass" refers to classification where each event is identified as one class among three or more possible classes [20, 21].

This study aimed to address these challenges by proposing a novel framework for arrhythmia detection and classification within a CDSS. The proposed framework integrates both local and global features to capture intermittent changes in ECG signals effectively. Additionally, it enables multiple classifications from a single input without extensive inferences using a short sliding window approach. This design enhances the framework's capacity to capture temporal dependencies by fusing local and global temporal patterns. Our proposed framework comprises an encoder-decoder architecture [22, 23, 24], a temporal convolution network (TCN) [25] with multiscale temporal information fusion (TIF), a dilated convolution layer [26], and a multihead self-attention (MHA) mechanism [27]. For the baseline architecture, we utilized LinkNet [22], a semantic segmentation network that employs encoder-decoder structures, skip connections, and residual blocks. Furthermore, we proposed a combined loss function, incorporating categorical cross-entropy (CCE) loss and multiple weighted Dice loss, to address class imbalances. The MIT-BIH arrhythmia database (MITDB) [28] and the MIT-BIH AFIB database (AFDB) [28] were utilized in this study. Additionally, a novel fusion approach was proposed to integrate global flow information via multiscale TIF and local temporal interactions through the TCN block, addressing the challenge of simultaneously detecting positional and contextual information. We further introduced a fusion strategy that combines temporal MHA in the encoder with upsampled features in the decoder. This integration adds features that emphasize the interaction of temporal information through the MHA, which are then combined with the features restored by the decoder to enhance the richness and detail of contextual information.

The following is a summary of the study's main novelty. We proposed a multiscale TIF block composed of large kernels in both series and parallel structures to capture long-range arrhythmic dependencies from various perspectives and detect changes in local and global patterns. Multiscale TIF blocks are efficiently combined with TCN, which significantly improves the detection and classification performance for the onset and ending point of multiple arrhythmias, including paroxysmal arrhythmias, without relying on overlapping sliding window techniques. Our proposed framework introduced an input module utilizing the longest nonoverlapping sliding window approach. This design allows for the simultaneous detection and classification of arrhythmia onset and ending points while capturing local and global cardiac dynamics from arrhythmic flows. As a result, the proposed framework employs a computationally efficient input shift strategy. The new fusion strategy was devised to integrate the feature maps extracted by the encoder with the information obtained from the multiscale TIF in the decoder. Consequently, the proposed framework demonstrates significant improvements over previously developed models in terms of duration and episode performance.

# 2 Materials and Methods

## 2.1 Dataset

The experiments were conducted using three ECG datasets, summarized in Table I. A detailed description including characteristics are as follows. First, MITDB presents that two-channel ambulatory ECG data were collected from 47 patients at the MIT-BIH Arrhythmia Laboratory to form this database. Each recording lasted for approximately 30-min. The sampling rate was 360 Hz. The recordings in the database correlate with 14 different types of arrhythmia and each exhibiting a range of clinically significant arrhythmias [29]. In this study, we chose to focus on 10 arrhythmic rhythm types out of the total of 14 since they had high levels of arrhythmia prevalence. The remaining classes including atrial bigeminy, ventricular bigeminy, ventricular trigeminy, idioventricular rhythm, and 2-degree heart block were labeled as "other", and 10 classes were categorized using the MITDB. Secondly, AFDB have characteristics that two-channel ECG records for 25 patients with atrial fibrillation (AFIB) with paroxysmal symptoms are available in the AFDB. Each data file lasts 10-hour. A total of 250 samples per second were taken when sampling the signals. AFIB, atrial flutter (AFL), atrioventricular rhythm (AVR), and normal sinus rhythm (NSR) are all included in the database [2]. Unlike the MITDB, we used all 4 classes, including AFIB, AFL, AVR, and NSR. MADB showed that two-channel ECG records for 22 patients who experienced episodes of sustained ventricular tachycardia (VT) and ventricular fibrillation (VFIB) in the MADB. MADB has been shown to exhibit characteristics of arrhythmia and signal noise, which have been observed in individuals with a history of malignant arrhythmias. Each data file lasts 35-min. A total of 250 samples per second were taken when sampling the signals [29]. In this study, we internally tested the 8-class classification performance including arrhythmia and noise in MADB using 5-fold cross validation. In addition, we chose to focus on 8 arrhythmic rhythm types out of the total of since they had high levels of arrhythmia prevalence. The remaining classes including ventricular bigeminy, ventricular trigeminy, idioventricular rhythm, 1-degree heart block, asystole, ventricular escape rhythm, and high grade ventricular ectopic activity were labeled as "Other", and 8 classes were categorized using the MADB. Next, we chose AFIB, AVR, and NSR to check generalization ability. The remaining classes including VT and VFIB were removed to compare with the generalization performance of the AFIB, AFL, and NSR classes tested with AFDB.

Table 1. Summary of three utilized databases.

| Database | No. of Subject | Sampling Rate (Hz) | Types of Arrhythmia | No. of Sample |
|---|---|---|---|---|
| MITDB | 47 | 360 | AFIB, AFL, AVR, SVT, VT, VF, PREX, SBR, NSR, Other | 13364 |
| AFDB | 25 | 250 | AFIB, AFL, AVR, NSR | 74838 |
| MADB | 22 | 250 | AFIB, AVR, SVT, VT, VFIB, Noise, NSR, Other | 9461 |

The primary reasons for selecting these three datasets were as follows: First, the MITDB, AFDB, and MIT-BIH malignant arrhythmia database (MADB) contain multiple and multiclass arrhythmias. These datasets include data with durations exceeding 30 minutes or 10 hours, allowing for the creation of input sequences where various arrhythmias may or may not be present, depending on the window segmentation. In contrast, many other ECG datasets consist only of pre-segmented input data with corresponding arrhythmia labels. Furthermore, these three datasets are frequently used in arrhythmia classification studies and include a variety of arrhythmias, such as AFIB, AFL, VT, and SVT, as detailed in Tables 2 and 3. This diversity made it suitable to test the model's performance in scenarios involving multiple arrhythmias. Although other long-term ECG databases exist, many contain only normal rhythms or binary classifications, such as AFIB and non-AFIB. Therefore, the performance of the model was evaluated in solving one-input/multiple and multiclass output problems using only these three datasets. The proposed framework was applied to each dataset for performance evaluation, and the MITDB was specifically used for the ablation study.

Table 2. Detailed Information of Arrhythmia Episode and Presented Type in the MITDB, AFDB, and MADB.

| Database | Subjects with Arrhythmia | Num. of Average Episode | Num. of Average Type | Shortest | Longest |
|---|---|---|---|---|---|
| MITDB | 28/47 | 44.4±35.8 | 2.9±2.5 | VT (1.5 s) | AFIB (378.3 s) |
| AFDB | 25/25 | 23.7±27.5 | 2.4±0.7 | AVR (1.5 s) | AFIB (37142.0 s) |
| MADB | 22/22 | 13.9±15.9 | 2.5±1.7 | VFIB (0.7 s) | Other (1664.7 s) |

Table 3. Detailed information of arrhythmia percentage of events occuring in the MITDB, AFDB, and MADB.

| Database | AFIB (%) | AFL (%) | AVR (%) | SVT (%) | SBR (%) | PREX (%) | VT (%) | VFIB (%) | NSR (%) | Noise (%) | Other (%) |
|---|---|---|---|---|---|---|---|---|---|---|---|
| MITDB | 0.99 ±4.30 | 0.27 ±0.96 | 0.18 ±0.94 | 0.11 ±0.73 | 2.27 ±14.90 | 0.93 ±6.11 | 0.27 ±0.92 | 0.18 ±1.19 | 79.52 ±79.05 | – | 15.27 ±11.53 |
| AFDB | 43.47 ±37.15 | 0.17 ±0.36 | 0.06 ±0.16 | – | – | – | – | – | 56.30 ±37.12 | – | – |
| MADB | 7.02 ±22.88 | – | 4.00 ±16.47 | 3.42 ±12.21 | – | – | 10.97 ±13.49 | 6.82 ±8.05 | 39.45 ±30.54 | 13.33 ±21.99 | 14.54 ±21.63 |

The 14-class of MITDB and MADB were utilized for the analysis, with the MITDB classes being reset to a range of 10 and the MADB classes being reset to a range of 8. In the present study, the analysis was conducted using all 4-class arrhythmic rhythms. Furthermore, in contrast to prior studies that evaluated generalization capability employing two categories of AFIB and non-AFIB [30], our approach entailed the verification of 4-class AFDB and 3-class MITDB utilizing MITDB-based and MADB-based models.

## 2.2 Data Pre-processing

ECG leads V1, V5, and II are each represented by two records in the MITDB. Typically, lead II is used to detect arrhythmia [14]. Because it is typically used for all patients, we only used lead II in the ECG signals. ECG data were resampled at 256 Hz. Subsequently, a third-order Butterworth filter with three orders was used to adjust for the baseline wandering. The ECG segment labels in the MITDB

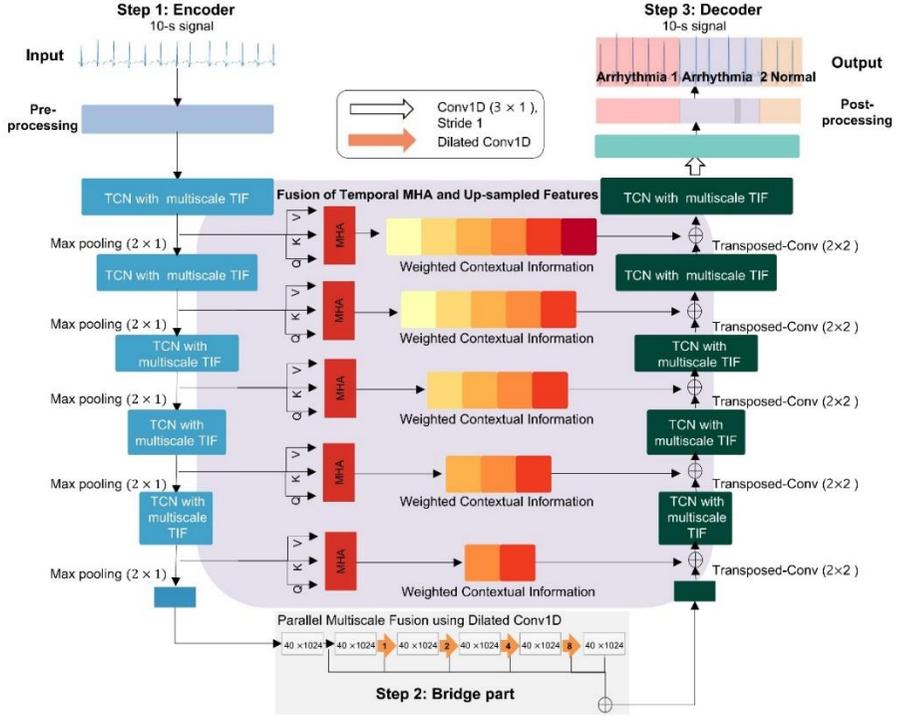

Figure 2. Overview of our arrhythmia detection and classification framework. Our suggested framework is composed of three steps. Step 1 encodes information by compressing features in the input data. The normalized data are then inserted into five TCN blocks with a multiscale TIF. The output of each TCN block is passed to the corresponding decoder after applying the MHA. The weighted contextual information passed to the decoder fuses the upsampled information in TCN with the multiscale TIF information of the decoder. Next, the information derived from the last residual block is fed to the parallel multiscale block using dilated Conv1D and then applied in Step 2. The bridge part includes a dilated Conv1D of the shortcut connections to form a parallel multiscale fusion structure for more comprehensive understanding the meaning of the arrhythmia information. Step 3 is composed of five TCN blocks with multiscale TIF and a transposed convolution layer. Temporal convolutional network: TCN. Temporal information fusion: TIF. One-dimensional dilated causal convolution: dilated Conv1D. Multihead self-attention: MHA. Queries: $Q$. Keys: $K$. Values: $V$.

represent 2560 samples, each with 10-s recordings. Each sample was assigned a class label.

ECG segment values were not used because each ECG data point had dissimilar amplitude scaling and vanishing offset effects. To eliminate these effects, we employed normalization, which involves scaling the signals to the same level. The Z score normalization method was used to normalize each ECG segment. AFDB and MADB were subjected to an identical set of preprocessing steps to those used for the MITDB.

## 2.3 Proposed Framework

The proposed framework comprises three main components, as shown in Figure 2: an encoder, a bridge part, and a decoder. Step 1 involves encoding information by compressing the features in the input data. The input to the encoder is a 10-second signal. Preprocessed and normalized data are fed into the encoder to facilitate feature extraction, with the preprocessing procedure. The encoder consists of five TCN blocks [31] paired with a multiscale TIF block. The TCN blocks are designed to capture local long-term sequential patterns, while the multiscale TIF block is configured to extract ECG signal patterns globally. These extracted features are then summed. The output from each TCN block is transferred to the corresponding decoder after the application of the MHA mechanism. This weighted contextual information is passed to the decoder, which integrates the upsampled information from the TCN with the multiscale TIF features processed by the decoder. The bridge part incorporates dilated convolution and shortcut connections to provide a more comprehensive understanding of arrhythmia-related information. Similar to the encoder, step 3 comprises five TCN blocks with multiscale TIF and a transposed convolution layer. These layers enhance the representation capability of the network. After completing these steps, the compound loss function is computed using the weighted Dice coefficient and the categorical cross-entropy (CCE) loss [32]. Finally, postprocessing is applied to correct any incorrect output samples of less than 256 samples within the arrhythmia 2 periods, despite the proposed method outputting different rhythms, including arrhythmia 1, arrhythmia 2, and normal rhythms.

### 2.3.1 Local-Global Temporal Feature Fusion Component

This component consists of a TCN block for local temporal feature extraction and a multiscale TIF block for global pattern feature extraction. After extracting local temporal and global pattern features, this component fuses the extracted features to enhance representation. The input of this part is the processed ECG signal, which is a two-dimensional matrix with determined dimensions (batch size, 2,560).

TCNs are significantly smaller than LSTM models, making them well-suited for sequence modeling tasks with an extended effective history. For this reason, we employed the input data as a TCN with 10-second signals of size 2560 × 1 and processed the data at various time intervals to capture local temporal dynamics. The TCN architecture includes two layers of dilated causal convolutions (dilated Conv1D), interspersed with layer normalization [33], a rectified linear unit (ReLU) activation function [34], and a drop-out layer [35]. The dilated Conv1D [26] enables an exponentially large receptive field, which is advantageous for capturing long-range dependencies in the data. Given a time sequence of input $E \in R^n$ and filter $f : \{0, \cdots, k-1\} \to R$, the dilated Conv1D operation $T$ implemented on the element $t$ of a time step $E$ is defined as:

$$T(E_t) = (E \cdot d)(t) = \sum_{i=0}^{k-1} f(i) \cdot E_{t-d \cdot i}, E \leq 0 := 0$$

$$Z = (T(E_1), T(E_2), \cdots, T(E_z))$$

$d$ is the dilation rate, $k$ is the kernel size, $t - d \cdot i$ counts the number of connections from previous nodes to the current node and $z$ is the output sequence with a length of $Z$

The multiscale TIF block was inspired by Ibtehaz et al. [36] and Peng et al. [37]. This block was composed of a series of three $10 \times 1$ convolution filters, other than a local kernel in the previous study [36. 38]. In addition, the output of the second and third $10 \times 1$ convolution filters was concatenated and passed through $1 \times 1$ convolution filters to efficiently connect. The sequence of large kernels with $10 \times 1$ filters enabled dense connections between the arrhythmic occurrence locations on the feature map for generating temporally stepwise semantic labels. Thus, we obtained multiscale temporal representations with three distinct temporal receptive fields, *i,e.,* $10 \times 1$, $12 \times 1$, and $14 \times 1$. Additionally, using different large kernels enabled the extraction of useful information without incurring information losses by inspecting points of interest at various scales. This approach improved the comprehension of global arrhythmic patterns. Consequently, the multiscale TIF block provided global arrhythmia classification and localization information. Subsequently, the $1 \times 1$ convolution on the multiscale TIF block facilitates efficient and comprehensive connectivity with the TCN, thereby enhancing comprehension of local-global arrhythmic patterns. Figure 3 illustrates the architecture of the multiscale TIF block.

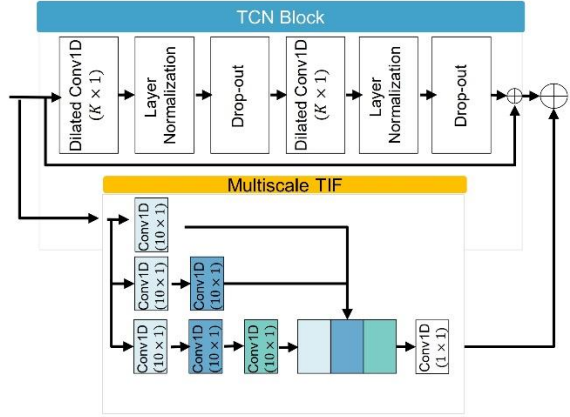

Figure 3. Architecture of the local-global temporal feature fusion component as combining TCN with multiscale TIF block. The multiscale TIF consists of three Conv1Ds. Each Conv1D conducts $10 \times 1$ convolutions. One-dimensional dilated causal convolutional layer. Dilated Conv1D. Temporal information fusion. TIF.

The output of every TCN with multiscale TIF block is activated by the ReLu non-linear function and then passed to the max-pooling layer for dimension reduction. The first block takes the processed 10-second ECG signal as input. The number of channels was set to 64, 128, 256, and 512, respectively. In the max-pooling layer, we took a stride of 2. Ultimately, the encoder step as a feature extraction has an output $1280 \times 64$, $640 \times 128$, $320 \times 256$, and $160 \times 512$, respectively.

### 2.3.2 MHA Component

The input sequences interacted and were combined in a weighted manner across all time steps using the self-attention mechanism [47]. The temporal MHA component was employed to learn contextual representations of arrhythmia information by capturing temporal interactions. The MHA was applied after the first through fifth TCN blocks, each paired with a multiscale TIF block. This mechanism enabled the model to focus on significant and relevant time steps over less important ones in the sequential feature maps during recognition. The output from the MHA was then integrated by utilizing the TCN with a multiscale TIF block in the decoder.

### 2.3.3 Bridge Part

Similar to the multiscale TIF block, this component utilized comprehensive multiscale temporal semantics by connecting dilated Conv1D components in parallel to expand the receptive field series. This part was inspired by the bridge component in D-LinkNet [48]. To achieve multiscale feature fusion, data from different branches were combined in parallel using a variety of convolutional architectures with varying dilation ratios. In the proposed method's bridge part, dilated convolution layers with dilation rates of 1, 2, 4, and 8 were employed. Consequently, the feature map extracted from the final layer of the encoder was utilized by the bridge part. This structure allowed the model to aggregate multiscale contextual information without significant loss [41, 42]. The bridge part processed 40 feature points of the feature map delivered through the TCN with the multiscale TIF block.

### 2.3.4 Decoder for Multiple- and Multi-class Classification

In this part, we constructed five TCNs with multiscale TIF blocks and five transposed convolutions [43]. The feature map $D_{in}$ after executing the bridge part, which was a transposed convolution layer, could be used to produce an upsampled output $D_{out}$ as follows:

$$D_1 = D_{in} \circledast k_1$$

$$D_2 = D_{in} \circledast k_2$$

$$D_3 = D_{in} \circledast k_3$$

$$D_4 = D_{in} \circledast k_4$$

$$D_{out} = D_1 \oplus D_2 \oplus D_3 \oplus D_4$$

Symbol $\circledast$ denotes a convolutional operation, $\oplus$ denotes periodical shuffling, $k$ is a kernel, and $k_1$ for $i = 1, \cdots, 4$.

After the transposed convolution, upsampled outputs and feature maps that passed through the temporal MHA were output to the TCN with the multiscale TIF block of the encoder and added. Then, the added feature map was passed through another TCN block with a multiscale TIF block to better understand the decoded contextual temporal information. The TCN with a multiscale TIF block operates similarly to those in the encoder. The decoder first executed a transposed convolution, added the upsampled outputs, obtained a feature map from a temporal MHA mechanism, and finally passed the results through a TCN block with a multiscale TIF block. This sequence was repeated five times. Finally, the outputs were fed to a Conv1D layer after passing through a transposed convolution layer.

### 2.3.5  Class-wise Loss Function and Optimization

The Dice loss function [44] was used and modified as it is widely recognized for medical segmentation tasks [44]. Although the Dice loss function addresses class imbalance to some extent, its capability to fully manage imbalance in classification tasks remains limited [45]. To complement this, CCE was utilized as an additional loss function because it evaluates class predictions at each time step individually and averages the results across all samples. Consequently, we combined the modified Dice loss function with CCE to suit the multiclass classification scenario and assigned different weights to each class, giving larger weights to less frequent classes to mitigate class imbalance. This loss function is expressed as follows:

$$CCE = -\sum_{i=1}^{n} y_{i,j} \log(p_{i,j})$$

$$L_{WDice} = 1 - \frac{2 \sum_{i=1}^{n} \sum_{j=1}^{c} w_{i,j}\, y_{i,j} p_{i,j}}{\sum_{i=1}^{n} \sum_{j=1}^{c} (y_{i,j} p_{i,j}) \times w_{i,j}}$$

$i \in n$ is sample obtained in the $i-th$ time step within the set of all time step samples $n$. $j \in c$ is the $j-th$ class in the set of arrhythmias $c$. $p_{i,j} \in P$, $y_{i,j} \in Y$, and $w_{i,j}$ are the predicted probability, ground truth, and weight for class $j$ regarding the $i-th$ time step sample. The target weight $w_{i,j}$ is computed as follows:

$$w_{i,j} = \frac{x^{tot}}{n \times x^j}$$

$x^{tot}$ is the total number of time step samples, and $n$ is the number of classes. $x^j$ is the number of time step samples in class $j$.

### 2.3.6  Postprocessing Procedure

The postprocessing procedure was as follows. When arrhythmias before and after samples with lengths below 256 were the same arrhythmia event, they were merged into a single event. For example, if AFIB was observed before and after a sequence of less than 256 samples, the result of this 10-s input was output as one AFIB event. However, if the events before and after an event containing fewer than 256 samples differed, the event with fewer than 256 samples was merged to the longer side among the occurrence times of the previous and subsequent events. For example, sinus bradycardia (SBR) occurred from 0 to 7-s and then changed to an AFIB event with less than 256 samples; in data where VT occurred from 7.8 to 10-s, the AFIB event was corrected to SBR.

### 2.3.7  Implementation Details

Implementation details, including optimal parameters and the computation environment for the proposed framework, are provided in Table 4. Table 4 presents details of the implementation of the proposed framework. The batch size was set to 64 and the starting learning rate was set to 0.00005 using the learning rate decay technique. To update the weights, we used an adaptive moment estimation (Adam) optimizer [46]. The raw ECG data were loaded into MATLAB R2020b for preprocessing, and then Python 3.7 was utilized to implement the model with Keras v1.3.1. An NVIDIA RTX 8000 GPU was used for training and testing. Specifically, the learning rate was set to 0.00005, the loss function was defined as a combination of multiclass weighted Dice loss with CCE, the number of epochs was set to 20, and the batch size was 64. The Adam Optimizer was used for weight updates. The convolutional kernel size in the TCN was set to 2, while the kernel size in the multiscale TIF block was set to 10.

Table 4. Detailed information of arrhythmia episode and presented type in the MITDB and AFDB.

| Hyperparameters | Value |
| --- | --- |
| Learning rate | 0.00005 |
| Optimizer | Adam [1] |
| Loss function | Dice loss with categorical cross entropy |
| Class balancing | Nonuniform (Class-wise weight)<br>• MITDB<br>AFIB: 1.0, AFL:19.1, AVR: 42.9, SVT: 4.4, SBR: 10.9, PREX: 43.8, VT: 10.6.4, VFIB: 43.8, NSR: 0.7, Other: 43.8<br>• AFDB<br>AFIB: 0.5, AFL: 3.2, AVR: 90.2, NSR: 0.5 |
| Pre-processing | Baseline wandering and min-max normalization |
| Batch size | 64 |
| Training epochs | 20 |

## 2.4 Comparison Models

To validate (i) the performance of the proposed framework and (ii) its training and inference time, we compared the proposed model with UNet [47], Res-UNet [48], LinkNet [22], and TCN-LinkNet (our baseline framework). We chose UNet [47], Res-UNet [48], and LinkNet [22] for the comparison since these frameworks are widely used in the biomedical field. TCN-LinkNet indicates that the multiscale TIF, temporal MHA, and bridge parts of our proposed model were removed, and only TCNs were included in the encoder and decoder of LinkNet. The same pre- and postprocessing steps were applied to the comparison models on the same dataset, consistent with the loss function described earlier.

A detailed explanation of the comparison models is provided as follows. UNet [47] comprises five compression stages that use a 1D convolutional layer and five upsampling states. The compressed feature maps in each compression stage concatenate the up-sampled feature maps using skip connections. The skipped connections recover the information lost during the compression stage. Therefore, UNet [47] is frequently used as a biosignal [3]. Res-UNet [48] uses five residual blocks as encoders and four upsampling layers as decoders. The encoder in Res-UNet [48] compresses the features from the previous encoder block. The decoder in Res-UNet [47] consists of five upsampling layers that combine compressed and regular feature maps through skip connections. This method applies various time series segmentation approaches [22]. Unlike Res-UNet [47], LinkNet [22] combines features through addition rather than concatenation. LinkNet [22] is an overall more efficient network because its decoder shares knowledge using fewer parameters. Therefore, this method is applicable to the field of biosignals. Lastly, TCN-LinkNet is an ablation framework used to test the multiscale temporal information fusion (TIF) block in the TCN block, bridge part, and MHA in the input of each encoder layer of the proposed method. Therefore, TCN-LinkNet comprises five TCN blocks as the encoder and five TCN blocks as the decoder.

## 2.5 Interpretation Using Grad-Cam

To highlight the regions relevant to an arrhythmia prediction given a time step label [49], gradient-weighted class activation mapping (Grad-CAM) was employed. This technique enhanced the interpretability of the model's predictions, allowing us to verify whether the labels assigned by the model were based on informative features [50]. To implement Grad-CAM in the proposed model, gradients of the final stack of filters in each network were computed for the relevant prediction class. These gradients allocated the significance of specific time steps for label prediction. By performing global average pooling on the gradients in each filter, filter importance weights were constructed, highlighting the filters whose gradients indicated a contribution to the class prediction. To generate a Grad-CAM heatmap, each filter in the convolution layer of the final decoder was multiplied by its corresponding importance weight and aggregated across all filters. The resulting heatmaps were then overlaid onto the original ECG images.

## 2.6 Statistical Analysis

Kruskal–Wallis tests were employed to compare the performance metrics of the frameworks with and without each proposed component, followed by post hoc analysis using pairwise Wilcoxon tests with the least-significant difference (LSD) technique [51]. The Kruskal–Wallis test and the Wilcoxon test with LSD are commonly used to identify significant differences among groups [52]. Additionally, the differences between various methods, including UNet [47], Res-UNet [48], LinkNet [22], the baseline framework of our proposed method, and the full proposed method, were evaluated using Kruskal–Wallis tests with the same post hoc method. The inference time differences among the proposed method and comparison models, such as UNet [47], Res–UNet [48], LinkNet [22], and TCN–LinkNet, were also assessed using Kruskal–Wallis tests, with pairwise Wilcoxon tests with LSD applied for post hoc analysis. A significance level of 5% ($p < 0.05$) was set for all statistical analyses.

# 3  Results

## 3.1  Ablation Study

### 3.1.1  Multiscale TIF

Table 5 lists the overall duration- and episode-based precision, recall, and F1-score for the proposed components. Adding the multiscale TIF block to the TCN block, which fuses features from local temporal information and global patterns, led to improvements in precision, recall, and F1-score for both duration- and episode-based performances. These metrics increased by more than 5% compared to the TCN block alone, as detailed in Table 5. Additionally, Figure 4 demonstrates that the frameworks incorporating the multiscale TIF block showed statistically significant performance enhancements compared to those without it.

Table 5. Results of an ablation study regarding the effect of each component using the MITDB.

| Bridge Block | MHA without Multiscale TIF Block | Multiscale TIF Block | Episode | | | Duration | | |
|---|---|---|---|---|---|---|---|---|
| | | | Precision (%) | Recall (%) | F1-score (%) | Precision (%) | Recall (%) | F1-score (%) |
| | | | 69.64± 2.53 | 87.74± 3.61 | 76.95± 2.78 | 94.24± 0.84 | 84.15± 3.88 | 88.70± 2.65 |
| √ | | | 69.50± 2.50 | 88.67± 3.78 | 77.23± 2.97 | 94.98± 1.31 | 85.34± 3.92 | 89.74± 2.82 |
| | √ | | 71.07± 2.48 | 93.54± 3.20 | 80.16± 2.59 | 97.90± 0.79 | 91.78± 3.66 | 94.65± 2.41 |
| | | √ | 71.00± 2.29 | 94.25± 2.75 | 80.36± 2.42 | 98.00± 0.72 | 92.80± 3.11 | 95.28± 2.03 |
| √ | | √ | 72.87± 2.78 | 94.41± 2.64 | 81.58± 2.45 | 98.54± 0.40 | 93.64± 2.68 | 95.96± 1.64 |
| | √ | √ | 73.08± 2.55 | 94.95± 2.56 | 81.93± 2.35 | 98.71± 0.39 | 94.23± 2.63 | 96.37± 1.60 |
| √ | √ | √ | 73.18± 2.53 | 95.12± 2.68 | 82.05± 2.38 | 98.79± 0.31 | 94.33± 2.82 | 96.45± 1.67 |

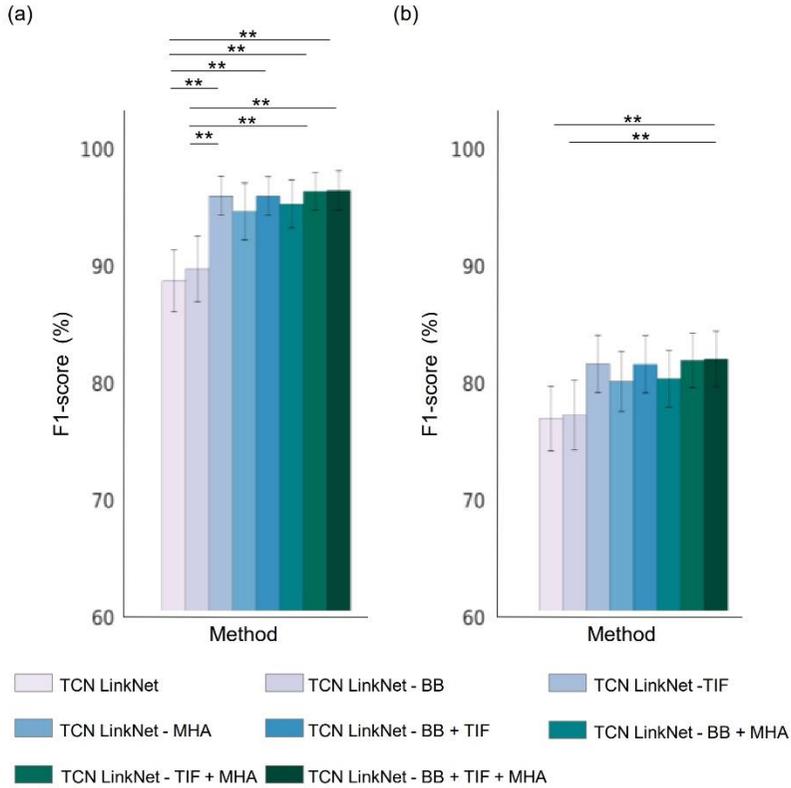

Figure 4. Statistical comparison among the overall F1-score produced by different components on the MITDB. (a) Overall duration F1-score. (b) Overall episode F1-score. The results are shown as overall means and ± standard deviations. ** $p < 0.01$ (the Kruskal-Wallis test with the LSD was used for post hoc analyses).

### 3.1.2 MHA Mechanism

We tested the effect of the MHA mechanism by removing the multiscale TIF block and bridge part (not the MHA). Table 5 shows that the MHA obtained precision, recall, and F1-score values of 71.07%, 93.54%, and 80.16%, respectively, in the episode classification test. In addition, the precision, recall, and F1-scores achieved in the duration classification test were 97.90%, classification performances of 69.64%, 87.74%, and 76.95% and 94.24%, 84.15%, and 88.70%, respectively, for both the duration and episode classifications. The MHA module achieved higher values for all metrics than those of the baseline for both duration and episode classifications; however, the MHA module did not statistically differ from the baseline, as shown in Figure 4.

### 3.1.3 Bridge Part

Table 5 indicates that without the multiscale TIF block in the TCN block, the MHA in the output derived from the TCN block with TIF and the bridge part, TCN-LinkNet achieves precision, recall, and F1- score values of 69.64%, 87.74%, and 76.95%, respectively, for episode-based classification. In addition, when the bridge part was added, the multiscale TIF block obtained precision, recall, and F1- score values of 69.50%, 88.67%, and 77.23%, 91.78%, and 94.65%, respectively. Without the MHA or multiscale TIF block, our baseline framework yielded lower respectively, for episode-based classification. The bridge part yielded a performance improvement of less than 1% over the baseline. However, when combined with the multiscale TIF block or the MHA, a performance improvement of more than 1% was obtained.

### 3.1.4 Postprocessing

The effect of postprocessing was evaluated using the MITDB. Table 6 shows that the proposed method with postprocessing achieved overall F1-scores of 96.45±1.67% and 82.05±2.38% for duration and episode classification, respectively. In contrast, the overall F1-scores for the duration and episode

Table 6. Comparison of effect with and without postprocessing using the MITDB.

| Method | F1-score (%) | |
| --- | --- | --- |
| | Duration | Episode |
| w/o Postprocessing | 96.60±1.63 | 49.61±4.19 |
| with Postprocessing | 96.45±1.67 | 82.05±2.38 |

classification without postprocessing were 96.60±1.63% and 49.61±4.19%, respectively.

## 3.2 Comparison with Other Models

### 3.2.1 Performance

We evaluated the performance of various models using 5-fold cross-validation on the 10-class MITDB, the 4-class AFDB, and the 8-class MADB to assess the effectiveness of the proposed method, as shown in Tables 7, 8, 9, and 11. The same test database was used for all models to ensure reasonable comparisons. Additionally, the combined loss function of our proposed method was applied consistently across all models.

Specifically, we compared the class-wise and overall duration and episode F1-scores with those of the comparative models to assess the performance of the proposed framework in class-wise arrhythmia classification, as shown in Table 7 and 8. The class-wise duration F1-scores of the proposed framework were 96.89±1.32, 96.04±1.22, 95.54±3.00, 95.54±3.37, 99.50±0.75, 95.90±1.21,

Table 7. Comparison among overall F1-score of the models using the MITDB, AFDB, and MADB.

| Method | MITDB (10-class) | | | AFDB (4-class) | | | MADB (8-class) | | |
|---|---|---|---|---|---|---|---|---|---|
| | Duration (%) | Episode (%) | Dice score (%) | Duration (%) | Episode (%) | Dice score (%) | Duration (%) | Episode (%) | Dice score (%) |
| UNet | 78.22±2.19 | 67.44±1.17 | 83.19±6.59 | 96.07±0.74 | 90.79±0.77 | 96.22±0.78 | 92.50±1.12 | 74.56±1.14 | 82.51±1.12 |
| Res-UNet | 73.82±3.70 | 63.84±2.83 | 59.99±15.49 | 91.24±1.20 | 87.28±1.22 | 90.85±1.59 | 57.14±0.84 | 53.84±0.32 | 57.14±0.84 |
| LinkNet | 69.21±3.31 | 63.34±2.12 | 47.87±19.75 | 93.54±1.26 | 89.11±0.67 | 71.19±1.58 | 50.58±0.29 | 48.74±0.69 | 50.58±0.29 |
| TCN-LinkNet | 88.70±2.65 | 76.95±2.78 | 89.39±2.56 | 96.73±1.90 | 97.11±0.84 | 95.70±1.24 | 83.70±0.68 | 78.08±0.49 | 83.87±0.64 |
| Proposed method | 96.45±1.67 | 82.05±2.38 | 96.31±1.25 | 97.57±1.11 | 98.31±0.76 | 97.45±1.46 | 86.16±0.61 | 78.98±0.37 | 86.06±0.73 |

96.17±3.06, 96.17±3.06, 96.68±1.17 and 95.93±1.33 for AFIB, AFL, AVR, SVT, SBR, PREX, VT, VFIB, NSR and Other, respectively. The class-wise and overall duration F1-scores achieved by the proposed framework were the highest compared to those of the other models. The class-wise episode F1-scores of the proposed framework were 80.72±1.25, 84.87±1.26, 88.55±5.57, 74.49±5.92, 99.72±0.56, 62.74±3.76, 86.54±3.57, 79.82±3.85, 75.48±2.23, and 82.25±1.38 for AFIB, AFL, AVR, SVT, SBR, PREX, VT, VFIB, NSR, and Other, respectively. When the class-wise performance of the episode F1-score was compared with the comparison models, all except AVR and PREX achieved the highest performance.

Table 8. Comparison among class-wise F1-score of the models using the MITDB.

| Method | Metrics | AFIB | AFL | AVR | SVT | SBR | PREX | VT | VFIB | NSR | Other |
|---|---|---|---|---|---|---|---|---|---|---|---|
| Unet | Duration (%) | 81.36 ±3.80 | 78.57 ±2.49 | 79.68 ±2.26 | 81.67 ±3.51 | 86.41 ±1.69 | 79.16 ±2.35 | 74.67 ±4.98 | 74.66 ±5.53 | 87.08 ±1.22 | 58.92 ±4.46 |
| Res–Unet | | 75.93 ±3.50 | 73.30 ±5.28 | 79.46 ±6.54 | 78.59 ±5.70 | 85.97 ±2.05 | 73.00 ±3.45 | 68.11 ±7.76 | 66.37 ±5.56 | 84.20 ±1.69 | 53.28 ±6.74 |
| LinkNet | | 72.53 ±4.55 | 66.96 ±3.33 | 68.36 ±5.94 | 76.58 ±4.70 | 76.18 ±2.21 | 70.64 ±4.10 | 63.95 ±8.32 | 66.34 ±5.67 | 81.79 ±2.02 | 48.73 ±4.40 |
| TCN–LinkNet | | 88.26 ±2.44 | 88.35 ±3.15 | 94.04 ±4.04 | 89.96 ±4.82 | 90.28 ±2.44 | 90.10 ±1.51 | 89.28 ±3.98 | 82.73 ±2.94 | 92.06 ±1.66 | 86.24 ±3.05 |
| Proposed method | | 96.89 ±1.32 | 96.04 ±1.22 | 95.54 ±3.00 | 95.54 ±3.37 | 99.50 ±0.75 | 95.90 ±1.21 | 96.17 ±3.06 | 96.17 ±3.06 | 96.68 ±1.17 | 95.93 ±1.33 |
| Unet | Episode (%) | 70.87 ±2.86 | 72.68 ±2.67 | 82.12 ±6.30 | 68.17 ±4.02 | 58.30 ±3.12 | 67.01 ±2.65 | 72.47 ±4.26 | 67.45 ±3.23 | 65.85 ±1.19 | 49.43 ±3.46 |
| Res–Unet | | 67.22 ±2.87 | 69.18 ±3.20 | 73.81 ±10.09 | 65.33 ±4.70 | 58.73 ±3.60 | 62.80 ±4.37 | 68.01 ±5.52 | 63.00 ±5.30 | 64.26 ±1.27 | 46.08 ±7.76 |
| LinkNet | | 67.13 ±2.71 | 66.27 ±4.01 | 83.58 ±5.70 | 65.04 ±3.43 | 51.77 ±3.45 | 61.67 ±4.86 | 63.50 ±6.08 | 67.60 ±6.46 | 63.69 ±1.47 | 43.10 ±3.35 |
| TCN–LinkNet | | 74.63 ±1.15 | 79.48 ±2.70 | 98.65 ±2.09 | 72.66 ±5.47 | 60.58 ±4.05 | 76.20 ±1.89 | 82.65 ±3.37 | 74.22 ±5.46 | 67.15 ±2.35 | 81.97 ±6.11 |
| Proposed method | | 80.72 ±1.25 | 84.87 ±1.26 | 88.55 ±5.57 | 74.49 ±5.92 | 99.72 ±0.56 | 62.74 ±3.76 | 86.54 ±3.57 | 79.82 ±3.85 | 75.48 ±2.23 | 82.25 ±1.38 |
| Unet | Dice score | 84.64 ±2.72 | 82.95 ±1.65 | 84.2 ±4.22 | 87.19 ±2.77 | 89.72 ±1.37 | 86.15 ±2.77 | 84.01 ±4.98 | 73.22 ±1.87 | 90.9 ±1.28 | 68.89 ±5.34 |
| Res–Unet | | 65.28 ±2.74 | 57.58 ±1.94 | 72.8 ±4.24 | 52.67 ±2.56 | 81.27 ±2 | 65.75 ±3.17 | 49.17 ±4.49 | 48.15 ±5.53 | 79.35 ±1.6 | 27.91 ±15.92 |
| LinkNet | | 65.58 ±3.18 | 57.17 ±3.33 | 65.44 ±17.48 | 37.41 ±2.86 | 63.37 ±6 | 46.72 ±3.85 | 37.12 ±1.85 | 36.5 ±8.62 | 68.26 ±1.82 | 1.17 ±1.47 |
| TCN–LinkNet | | 88.02 ±2.24 | 88.82 ±1.94 | 95.83 ±1.01 | 88.99 ±6.6 | 89.67 ±2.74 | 89.05 ±1.33 | 88.22 ±6.56 | 87.9 ±4.73 | 91.62 ±1.79 | 85.8 ±3.29 |
| Proposed method | | 96.89 ±0.89 | 95.57 ±0.87 | 99.67 ±0.66 | 95.11 ±3.41 | 96.31 ±1.15 | 96.13 ±0.99 | 95.29 ±3.83 | 95.6 ±2.09 | 96.73 ±1.03 | 95.84 ±2.06 |

Regarding the 4-class performance on the AFDB, we checked the class-wise and overall arrhythmia classification performance of AFDB using duration and episode F1-score, as shown in Table 7 and 9. The proposed framework achieved

Table 9. Comparison among class-wise F1-score of the models using the AFDB.

| Method | Metrics | AFIB | AFL | AVR | NSR |
|---|---|---|---|---|---|
| Unet | Duration (%) | 97.83±0.24 | 96.43±0.41 | 91.77±2.23 | 98.25±0.13 |
| Res-Unet | | 96.86±0.47 | 95.35±0.42 | 75.44±4.09 | 97.30±0.42 |
| LinkNet | | 96.85±0.54 | 94.93±0.48 | 84.99±4.04 | 97.37±0.50 |
| TCN-LinkNet | | 98.14±0.49 | 96.54±0.45 | 93.58±3.39 | 98.68±0.28 |
| Proposed method | | 98.45±0.72 | 96.65±0.50 | 96.01±2.91 | 99.08±0.48 |
| Unet | Episode (%) | 83.33±0.30 | 95.64±2.05 | 91.77±1.98 | 92.43±0.48 |
| Res-Unet | | 87.18±0.58 | 96.00±0.37 | 73.39±4.98 | 92.56±0.44 |
| LinkNet | | 83.21±0.91 | 93.02±2.15 | 87.07±3.6 | 93.16±0.47 |
| TCN-LinkNet | | 96.03±0.45 | 99.00±0.35 | 96.75±2.79 | 96.64±0.42 |
| Proposed method | | 98.39±0.74 | 99.11±0.28 | 97.10±1.92 | 98.66±0.50 |
| Unet | Dice score | 97.91±0.23 | 96.55±0.35 | 92.12±2.36 | 98.29±0.17 |
| Res-Unet | | 96.76±0.42 | 94.89±0.44 | 74.54±5.11 | 97.22±0.4 |
| LinkNet | | 94.64±1.43 | 93.77±0.58 | 1.43±2.87 | 94.92±1.47 |
| TCN-LinkNet | | 97.64±0.41 | 95.55±0.48 | 91.51±3.7 | 98.09±0.35 |
| Proposed method | | 98.57±0.56 | 96.74±0.51 | 96.41±2.53 | 99.08±0.42 |

class-wise duration F1-scores of 98.45±0.72, 96.65±0.50, 96.01±2.91, and 99.08±0.48 in AFIB, AFL, AVR, and NSR, respectively. The proposed framework achieved class-wise episode F1-scores of 98.39±0.74, 99.11±0.28, 97.10±1.92, and 98.66±0.50 in AFIB, AFL, AVR, and NSR, respectively. The proposed framework achieved higher performance than the models in duration and episode class-wise F1-score.

We evaluated the class-wise and overall arrhythmia classification performance of 8-class MADB using duration and episode F1-score, as shown in Table 8 and 10. The proposed framework achieved class-wise duration F1-scores of 97.13±2.09, 98.15±0.46, 85.55±0.65, 97.46±0.06. 95.99±0.21, 43.07±0.50, 97.99±0.81, and 98.09±0.09 in AFIB, AVR, SVT, VT, VFIB, Noise, NSR, and Other, respectively. The proposed framework achieved class-wise episode F1-scores of 98.32±5.26, 97.29±1.07, 97.14±3.19, 73.78±0.29. 76.81±0.72, 33.57±0.21, 86.88±1.72, and 95.17±0.42 in AFIB, AVR, SVT, VT, VFIB, Noise, NSR, and Other, respectively. The proposed framework achieved higher performance than the models in duration and episode class-wise F1-score.

The class-wise Dice score of the proposed framework was 96.89±1.32, 95.57±0.87, 99.67±0.66, 95.11±3.41, 96.31±1.15, 96.13±0.99, 95.29±3.83, 95.60±2.09, 96.70±1.03, and 95.84±2.06 for AFIB, AFL, AVR, SVT, SBR, PREX, VT, VFIB, NSR, and other, respectively. All class-wise duration F1-scores achieved the highest performance compared to other comparative models. Regarding on the 4-class performance on the AFDB, the proposed framework achieved class-wise Dice scores of 98.57±0.56, 96.74±0.51, 96.41±2.53, and 99.08±0.42 in AFIB, AFL, AVR, and NSR, respectively. Next, the class-wise Dice score of the proposed method on the 8-class MADB was 98.13±2.09, 98.15±0.46, 85.55±0.65, 97.46±0.06. 95.99±0.21, 43.07±0.50, 97.99±0.21, and 98.09±0.09 in AFIB, AVR, SVT, VT, VFIB, Noise, NSR, and Other, respectively.

Table 10. Comparison among class-wise F1-score of the models using the MADB.

| Method | Metrics | AFIB | AVR | SVT | VT | VFIB | Noise | NSR | Other |
|---|---|---|---|---|---|---|---|---|---|
| UNet | Duration (%) | 96.56±0.32 | 96.99±0.41 | 77.69±5.09 | 91.91±0.64 | 91.61±0.04 | 40.21±0.39 | 94.61±0.04 | 94.73±0.32 |
| Res-UNet | | 88.54±0.79 | 86.30±1.32 | 0.00±0.00 | 71.77±0.89 | 68.08±1.88 | 28.13±0.01 | 76.18±1.58 | 83.57±0.24 |
| LinkNet | | 81.81±3.42 | 78.87±1.14 | 0.00±0.00 | 66.99±0.03 | 53.75±2.13 | 22.05±0.41 | 63.70±3.30 | 77.70±0.61 |
| TCN-LinkNet | | 95.10±0.12 | 97.80±0.40 | 83.80±2.10 | 92.90±0.70 | 91.60±0.40 | 40.90±0.03 | 95.60±1.40 | 96.20±0.30 |
| Proposed method | | 97.13±2.09 | 98.15±0.46 | 85.55±0.65 | 97.46±0.06 | 95.99±0.21 | 43.07±0.50 | 97.99±0.81 | 98.09±0.09 |
| UNet | Episode (%) | 91.61±0.63 | 96.36±0.63 | 84.93±5.66 | 71.39±0.88 | 74.37±1.10 | 28.67±0.34 | 82.37±1.02 | 86.86±1.00 |
| Res-UNet | | 90.28±1.15 | 93.26±1.61 | 0.00±0.00 | 58.80±0.56 | 60.01±1.84 | 20.69±0.54 | 72.01±0.54 | 74.44±0.90 |
| LinkNet | | 83.99±4.17 | 84.11±2.75 | 0.00±0.00 | 55.63±1.13 | 52.26±3.36 | 16.47±0.41 | 62.56±2.36 | 70.29±0.25 |
| TCN-LinkNet | | 97.04±0.36 | 98.36±0.91 | 92.84±2.63 | 72.90±0.40 | 75.99±0.89 | 31.35±0.15 | 80.99±0.89 | 93.63±0.21 |
| Proposed method | | 98.32±5.26 | 97.29±1.07 | 97.14±3.19 | 73.78±0.29 | 76.81±0.72 | 33.57±0.21 | 86.88±1.72 | 95.17±0.42 |
| UNet | Dice score | 96.56±0.32 | 96.99±0.41 | 77.69±5.09 | 91.91±0.64 | 91.61±0.40 | 40.21±0.39 | 92.64±0.80 | 94.73±0.32 |
| Res-UNet | | 88.54±0.79 | 86.30±1.32 | 0.00±0.00 | 71.77±0.89 | 68.08±1.88 | 28.13±0.60 | 78.28±1.18 | 83.57±0.24 |
| LinkNet | | 81.81±3.42 | 78.87±1.14 | 0.00±0.00 | 66.99±0.30 | 53.75±2.13 | 22.05±0.41 | 66.75±4.13 | 77.70±0.61 |
| TCN-LinkNet | | 95.70±1.20 | 98.00±0.40 | 83.50±2.20 | 93.30±0.70 | 91.70±0.40 | 41.10±0.40 | 93.12±0.80 | 96.40±0.30 |
| Proposed method | | 98.13±2.09 | 98.15±0.46 | 85.55±0.65 | 97.46±0.06 | 95.99±0.21 | 43.07±0.50 | 97.99±0.21 | 98.09±0.09 |

Moreover, we conducted the Kruskal-Wallis test followed by the LSD post hoc test. Figure 5 illustrates the statistical analysis results for the comparison models, including UNet [47], Res-UNet [48], LinkNet [22], TCN-LinkNet (baseline), and the proposed method. The results demonstrated that the proposed framework achieved statistically higher performance in terms of episode and duration F1-scores ($p < 0.05$).

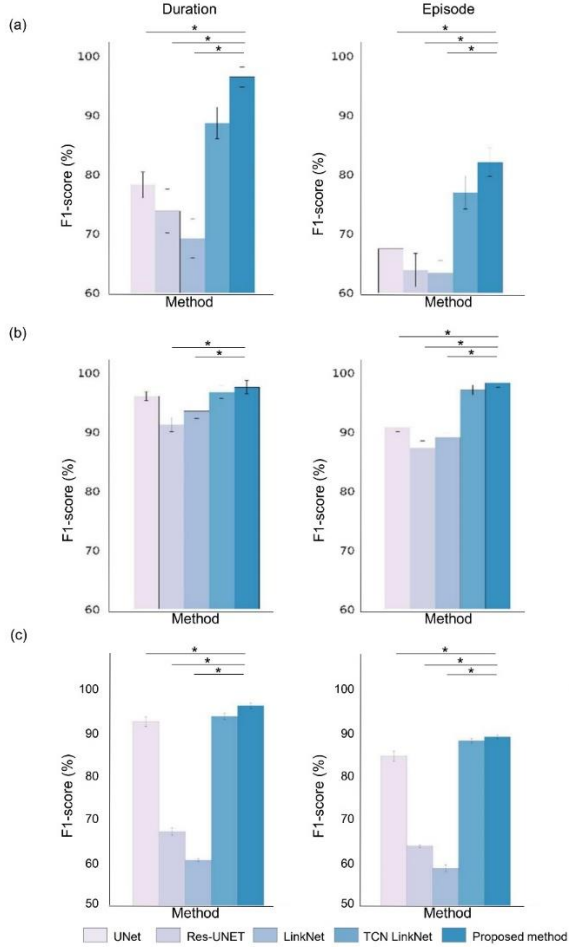

Figure 5. Statistical comparison among the overall F1-score performances of the various models, including UNet [47], Res-UNet [48], LinkNet [22], TCN-LinkNet (our baseline), and the proposed method. (a) Overall duration and episode F1-score achieved on the MITDB. (b) Overall duration and episode F1-score achieved on the AFDB. (c) Overall duration and episode F1-score achieved on the MADB. We compared the models on three databases (the MITDB, AFDB, and MADB). We only show the statistically significant differences between the proposed method and the comparison models. The results are presented as overall means $\pm$ standard deviations. $*$ $p < 0.05$ (the Kruskal-Wallis test with the LSD was used for post hoc analyses).

### 3.2.2 Generalization Ability

We tested the generalization ability using the cross-dataset method using three databases as shown in Table 11. First, we trained comparative models and proposed models using 4-class including AFIB, AFL, AVR, and NSR from MITDB, which contains various arrhythmia occurrence events. Afterward, common 4-class events were extracted from the AFDB, which was collected into groups representing atrial fibrillation characteristics, and the generalization ability for different group characteristics was tested. As a result, the proposed model achieved the highest performance in duration with an overall 36.29%. In particular, 19% and 27% higher performance than UNet was obtained in AFIB and AFL classes. In the episode, it obtained an overall score of 51.71%, achieving higher performance than the compared models. Compared to LinkNet, which showed the lowest overall performance, AFIB, AFL, and AVR showed high performance improvements of 17%, 76%, and 10%.

We performed a generalization ability test on MITDB, representing various arrhythmia populations after training using MADB collected from subjects with malignant arrhythmia characteristics. Since the three classes of AFIB, AVR, and NSR were commonly available in both databases, we extracted events for these three classes and used them for training and testing. As a result, we obtained the highest performance in terms of duration and overall performance of episodes compared to the comparative models. It achieved 9% higher performance in both AFIB and AVR than UNet, which showed the lowest performance in duration. episodes also showed 14% and 35% higher performance than UNet, which showed the lowest performance.

Table 11. Results of generalization ability comparison between the proposed model and different models.

| Method | Metrics | AFDB test after MTIDB training (4-class) | | | | | MITDB test after MADB training (3-class) | | | |
|---|---|---|---|---|---|---|---|---|---|---|
| | | AFIB | AFL | AVR | NSR | Overall | AFIB | AVR | NSR | Overall |
| UNet | Duration (%) | 23.74 | 9.76 | 0.00 | 62.52 | 24.00 | 4.71 | 0.05 | 89.73 | 31.49 |
| Res-UNet | | 24.67 | 0.20 | 0.11 | 63.92 | 22.22 | 4.24 | 1.94 | 88.83 | 31.67 |
| LinkNet | | 21.98 | 0.05 | 0.00 | 62.19 | 21.05 | 6.41 | 0.00 | 88.38 | 31.59 |
| TCN-LinkNet | | 41.57 | 31.51 | 0.89 | 57.30 | 32.82 | 10.85 | 2.46 | 83.77 | 32.36 |
| Proposed method | | 42.33 | 37.52 | 1.35 | 63.96 | 36.29 | 13.89 | 9.69 | 88.33 | 37.3 |
| UNet | Episode (%) | 21.35 | 10.01 | 0.00 | 88.73 | 30.02 | 4.28 | 0.00 | 95.66 | 33.31 |
| Res-UNet | | 22.49 | 0.14 | 0.41 | 89.88 | 28.23 | 4.33 | 25.00 | 94.28 | 41.20 |
| LinkNet | | 21.79 | 0.03 | 0.00 | 90.11 | 27.98 | 6.44 | 0.00 | 97.82 | 34.75 |
| TCN-LinkNet | | 36.6 | 43.41 | 10.15 | 63.59 | 38.44 | 10.84 | 16.67 | 88.91 | 38.8 |
| Proposed method | | 38.9 | 76.93 | 10.86 | 80.15 | 51.71 | 18.03 | 35.22 | 95.10 | 49.45 |

### 3.2.3 Training and Inference Time

We proposed method required longer training time per epoch compared to the other models due to its increased complexity. However, the inference time of the proposed method was not significantly different from that of the comparison models. Next, an analysis was conducted to assess the inference time of the proposed model. This analysis was based on measured ECG data from each subject in each database. Additionally, the inference time when analyzing real-world 24 hour Holter monitoring ECG data was considered. Table 12 shows the inference time of the proposed model based on measured ECGs of individual subjects for each database including MITDB, AFDB, and MADB. Table 13 indicates that the performance of the proposed model differed significantly from that of the comparison models ($p < 0.05$). A post hoc analysis showed that the proposed method was not statistically different from Res-LinkNet and TCN-LinkNet. However, the proposed method is statistically different from UNet and LinkNet.

Table 12. Comparison of training and inference time using the MITDB.

| Method | Time per Epoch (Sec) | Inference Time per Batch Size (Sec) |
|---|---|---|
| UNet | 43.40±0.31 | 0.11±0.01 |
| Res-UNet | 58.3±0.30 | 0.34±0.08 |
| LinkNet | 35.35±0.34 | 0.11±0.01 |
| TCN-LinkNet | 36.35±0.30 | 0.22±0.01 |
| Proposed method | 87.5±0.55 | 0.50±0.01 |

Table 13. Statistical comparison and multiple comparison of inference time using the MITDB.

| Kruskal-Wallis test | | Post-hoc | | | |
|---|---|---|---|---|---|
| | | Proposed Method -UNet | Proposed MEthod -Res-UNet | Proposed Method -LinkNet | Proposed Method -TCN-LinNet |
| Chi-Squre | 27.979 | 15.86 | -3.47 | 21.08 | -1.24 |
| p-value | <0.001 | <0.01- | 0.16- | <0.01- | 0.07- |

Table 14. Inference time of proposed model based on measured ECGs of individual subjects.

| Database | Duration measured | Time per Subject (Sec) |
|---|---|---|
| MITDB | 30 min | 1.55±0.08 |
| AFDB | 10 hour | 31.94±1.30 |
| MADB | 35 min | 1.80±0.10 |
| Holter Monitoring | 24 hour | 74.25±2.21 |

Table 14 shows the inference time of the proposed model based on measured ECGs of individual subjects for each database including MITDB, AFDB, and MADB. In addition, the inference time of ECG data collected by 24 hour Holter monitoring is also presented.

## 3.3  Clinical Interpretability

Figure 6 demonstrates the capability of the proposed method to extract temporal features associated with clinical patterns. For NSR, the heatmap images revealed that the proposed method focused on the P, R, and T waves. For arrhythmias, the proposed method highlighted irregular QRS complexes. Specifically, Grad-CAM emphasized an irregular complex, a T wave, and a fibrillation wave in the PREX and AFL cases. Additionally, when multiple arrhythmias were present, the proposed method identified and underscored the importance of the onset and ending points between consecutive arrhythmias.

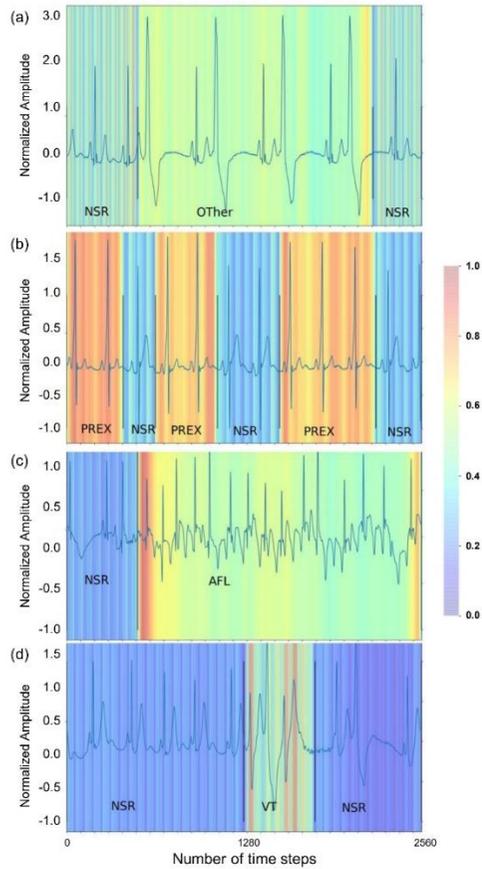

Figure 6. Heatmaps of the last convolution layer activations produced in the final decoder using Grad-CAM in the proposed method. (a) NSR and Other. (b) NSR and PREX. (c) NSR and AFL. (d) NSR and VT. The heatmaps of the activity in last convolution layer indicate that the proposed method distinguished multiple and multiclass arrhythmias. The color scale indicates the value of the time step. Our proposed method concentrated on aspects from blue to red and normalized them between 0 to 1. Red represents high arrhythmia significance, and blue signifies low arrhythmia significance for the proposed method.

# 4  Discussion

We proposed a computationally efficient multiclass arrhythmia classification system capable of learning both local and global patterns. To understand the factors contributing to its performance improvements, we conducted a series of ablation experiments. Furthermore, the proposed framework demonstrated superior performance compared to baseline models across the duration, episode, and Dice score metrics. We discuss the outcomes of this analysis from multiple perspectives as follows.

## 4.1  Model Component Analysis

We conducted ablation studies to evaluate the impact of three primary components: (i) a TCN block combined with a multiscale TIF block, (ii) an MHA mechanism applied to the output of the TCN block combined with a multiscale TIF block, and (iii) a TCN block combined with a multiscale TIF block. The findings of these ablation experiments are summarized in Figure 4.

We evaluated the effectiveness of the multiscale TIF block through an ablation study that involved removing and reinserting this component. The multiscale TIF block fuses different resolutions of global patterns using a multiscale large kernel, which primarily enables the capture of diverse long-range global temporal information [37, 53, 54]. This enhancement improves the model's ability to accurately identify the onset and ending points of arrhythmias. Consequently, $TP_{Duration}$ increased, resulting in statistically significant improvements in precision, recall, and F1-scores for duration-based evaluations compared to the framework without the multiscale TIF module. Notably, $FN_{Duration}$ decreased, and the recall metrics improved by approximately 6% or more over the baseline in both episode－ and duration－based evaluations. Additionally, $FP_{Duration}$ was reduced, leading to an approximate 1% increase in the precision metric in the duration－based evaluation.

From the results and analyses discussed above, the proposed multiscale TIF block demonstrates the ability to learn various global patterns by fusing different resolutions through a multiscale large kernel. This capability enhances the model′s ability to comprehend and detect contextual arrhythmia information at onset and ending points. In summary, the application of the multiscale TIF block has been shown to improve the overall accuracy of the model. These findings support the claim that the multiscale TIF block enhances the model's discriminatory ability, particularly concerning the onset and ending points of arrhythmia.

The MHA module contributes by emphasizing distinctive local temporal interaction features before passing them to the decoder, which allows the model to better learn and understand contextual information. Furthermore, when the MHA module was combined with the multiscale TIF block, all performance metrics were statistically improved compared to the baseline and the bridge component alone. This result indicates that the MHA module effectively captures distinct features within the fused long－range local and global temporal information and transmits them to the TCN block with the multiscale TIF block in the decode.

We analyzed the effect of postprocessing as shown in Table 7. These results indicated that the effect of postprocessing was similar in terms of the overall duration F1－score, and the overall episode F1－score differed between the cases with and without postprocessing. The episode performance was calculated using counts of true positives, false negatives, and false positives [55], whereas duration performance was measured based on the overlap between the ground truth periods and the periods classified by the model. In terms of duration, there was no significant difference in the overall F1-score with and without postprocessing due to similar overlapping periods between the ground truth and the model's classifications. However, the proposed method frequently outputs a fine region between arrhythmias, which was considered a false positive [55]. These false positives negatively affected the model's episode performance. Consequently, while postprocessing had no effect on duration performance, it did have an impact on episode performance. Finally, because the proposed method impacted episode performance, developing an advanced method for classifying fine regions between arrhythmias is recommended for future work.

We performed a comparative analysis between the baseline models and the proposed framework using 5-fold cross-validation on the 10-class MITDB, the 4-class AFDB, and 8-class MADB. We analyzed the overall and class-wise duration and episode F1-scores and Dice scores. The experimental results indicated that the proposed framework achieved statistically significant improvements over the comparison models. The reasons for these higher performances are as follows. First, TCN-LinkNet was designed for local temporal information extraction \cite{lea2017temporal}, enabling it to learn features that account for interactions between local time steps over long sequences. Consequently, the overall duration and episode F1-scores on the MITDB improved by 19% and 13%, respectively, compared with those of LinkNet. Additionally, the overall duration F1-score improved by 10% and the episode F1-score by 9% compared with those of UNet [47]. This improvement is attributed to TCN-LinkNet's ability to extract contextual information from interactions between local time steps, overcoming the limitation of UNet [47], which only utilizes local feature information through convolution filters [56]. Second, our proposed method fused local temporal interaction features and global pattern features by leveraging multiscale TIF to introduce global pattern features into the skip connections of TCN-LinkNet. Moreover, the proposed method incorporated the MHA framework to enhance feature representation further. TCN-LinkNet as our baseline model. Therefore, our proposed method successfully captured both local temporal information and various global temporal features, unlike UNet [47], Res-UNet [48], LinkNet [22]. Consequently, the duration and episode F1-scores increased by approximately 6% compared to those of TCN-LinkNet. A qualitative analysis was conducted on the failure cases of the compared models for single input-multiple output classification problems to explore the reasons why the proposed model solved these cases, as shown in Figure 7. Figure 7 shows an example of selected arrhythmia classification results between the actual class and the predicted class of the comparison model and the proposed model. Specifically, AFIB and AFL demonstrate nearly analogous patterns; however, the electrical impulses of AFL are organized, while those in AFIB are not. Furthermore, the electrical pattern of AVR that originates near or within the atrioventricular node, as opposed to the sinoatrial node, is challenging to discern from NSR. Consequently, in cases where NSR emerged initially, subsequently accompanied by paroxysmal AFIB for a transient period, and culminating in reversion to NSR, AFIB remained undetected and was erroneously classified as NSR. Moreover, there were instances where AFL was erroneously classified as AFIB and instances where the rare label AVR went undetected. TCN-LinkNet (baseline), serving as the foundational framework of the proposed approach, possesses the capacity to discern local temporal interaction patterns. The ability of TCN-LinkNet (baseline) to capture these interaction patterns enables the discernment and comprehension of these distinctions, thereby enhancing the classification performance of AFL. However, in scenarios where global patterns must be captured, such as in the PREX experiment where paroxysmal and normal patterns appear consecutively, TCN-LinkNet (baseline) outperforms other comparative models but falls short of the proposed framework.

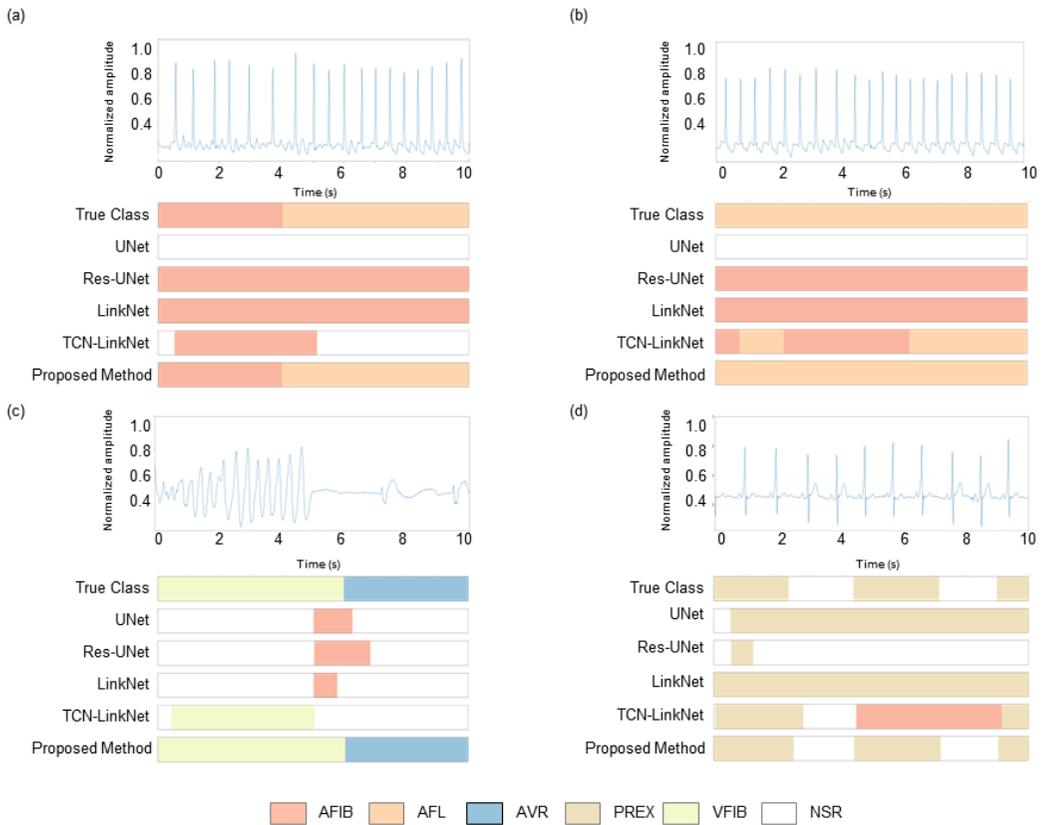

Figure 7. Qualitative comparison results of selected arrhythmia between true class and predicted class of baseline models and the proposed method on 10-class MITDB. (a) AFIB and AFL. (b) AFL. (c) VFIB and AVT. (d) PREX and NSR.

An evaluation was conducted to assess the performance of various arrhythmia classification and noise robustness metrics recorded in ECG measurements. The evaluation utilized 8-class MADB with inclusion of noise. The noise exhibited in MADB is attributed to electrode movement, and this type of noise manifests frequency characteristics analogous to those observed in VT and VFIB [29]. Consequently, the detection and classification of VT and VFIB, as well as noise in the input pattern, pose a significant challenge. Although the proposed framework and baseline models exhibit high false negative rates ($FN_{Duration}$ and number of $FN$), the proposed framework demonstrates enhanced arrhythmia detection/classification accuracy in noisy conditions compared to baseline models on the 8-class MADB. Consequently, these results indicated that the proposed framework effectively identifies arrhythmias in noisy environments.

We tested the generalization ability using ECG from groups with different arrhythmia characteristics and device inputs. MITDB is a group that includes uncommon but clinically important arrhythmias and is generally used to evaluate arrhythmia classification performance. AFDB mostly collected electrocardiograms from subjects with paroxysmal AFIB. Because the populations included in the two databases and the characteristics of the measuring devices are different, even the same arrhythmia label has unique characteristics for each patient. In particular, the number of occurrences of AVR was very low compared to AFIB and AFL, and the patterns were also diverse because it had various mechanisms of occurrence [57]. Therefore, detecting and classifying this class within the input pattern was a very difficult problem. However, the proposed framework showed higher detection rates and classification performance. The proposed framework showed performance improvement of about 2% and 27% in the AFIB and AFL classes in the AFDB test compared to TCN-LinkNet. Additionally, MADB tests showed performance improvements of approximately 8% and 18%. This indicates that the TIF module played a significant role in understanding the local-global arrhythmic pattern, and the MHA module and bridge module helped the overall classification performance.

The proposed framework showed the highest performance in both duration and overall performance of episodes. The improvement in generalization performance is due to understanding the information of the arrhythmia onset and ending point patterns through the proposed information fusion strategy using the multiscale TIF block and MHA, as described above. Therefore, the proposed model is proven to generally improve the accuracy of overall arrhythmia onset and ending points.

We analyzed the training and inference times of the proposed model on the MITDB and compared them with those of the alternative methods. Although the proposed method required longer training time per epoch than the alternatives, its inference time was not significantly different from that of the other methods. The time required to analyze ECG data in a clinical setting can be categorized into two distinct types: emergency monitoring and non-emergency diagnosis. The process of emergency monitoring necessitates the analysis of a singular input in real time, within a span of 100 milliseconds. Conversely, post-analysis systems, such as Holter monitoring, are required to analyze the complete data set within a span of 60 minutes [58]. The proposed framework is capable of analyzing 24 hours of measured ECG data in 1.3 minutes as shown in Table 15. Notwithstanding the potential for augmented latency occasioned by the accumulation of data in the memory, the proposed framework's inference time remains within the bounds of the acceptable computational time threshold for clinicians. While the proposed method can be effectively used to inform clinicians of clinical arrhythmia data in practical settings, a more lightweight approach would be beneficial.

The proposed framework demonstrated the capability to extract features from multiple classes and learn effectively even when the input included multiple arrhythmias. Additionally, our model outperformed state-of-the-art models in arrhythmia duration and episode classification, highlighting its suitability for real-world clinical applications involving computational healthcare systems.

## 4.2 Clinical Interpretability

In NSR analysis, the weight values were distributed evenly across the P, R, and T waves, with consistent intervals between each value. NSR is characterized by regular intervals anchored around the R peak, with clear delineations of the P and T waves [59]. It is hypothesized that the proposed framework demonstrated a high level of clinical relevance due to its ability to recognize these standard patterns. In the analysis of arrhythmias, the proposed method detected irregular QRS complexes, elevated T waves, and gear-shaped fibrillation waves [59]. These distinct patterns, differing from the NSR pattern, are key diagnostic indicators used by clinicians for conditions such as AFL, PREX, and VT. Additionally, when multiple arrhythmias co-occurred, the weight assigned to the onset and ending points of each arrhythmia was heightened. Thus, the clinical interpretability provided by the proposed framework supports the integration of its predictive results into clinical workflows [60].

## 4.3 Comparison of Recent Works

Table 15 provides a comprehensive performance comparison between the proposed method and existing approaches for arrhythmia classification. We describe each corresponding method and report their performances using the overall F1-score metric.

Multi-class classification poses a significant challenge due to the diversity of arrhythmia types and the inter-patient variability in arrhythmia patterns. Previous research has aimed to address this challenge by developing deep learning frameworks utilizing publicly available databases. These studies have included single-lead approaches for portable ECG measurement and 12-lead systems for diagnostic purposes.

In single-lead research using the MITDB, a prior study classified only six types of arrhythmias. However, subsequent research increased the number of arrhythmia types classified to between eight and ten while maintaining a classification performance of 90%. Studies utilizing the AFDB, which includes four types of arrhythmias, indicate that Faust et al. [61] were able to classify only two types of arrhythmias. Kim et al. [9] extended this to four arrhythmias and achieved a classification performance exceeding 90%. For 12-lead studies, six arrhythmia classification investigations were conducted, taking into account variable ECG measurement durations. However, the studies mentioned above were constrained by the assumption of a single input-single output framework, which limited their capacity to accurately classify paroxysmal arrhythmias when they appeared multiple times in real-world scenarios. Specifically, extant methodologies have demonstrated deficiencies in the effective learning of contextual pattern changes preceding and following epileptic pattern changes. Consequently, the classification performance of the four classes was found to be significantly lower than that of the proposed method. A qualitative analysis of the failure cases of the compared models for single-input-multiple-output classification problems is presented in Figure 7.

Table 15 Comparison of classification performance of the proposed framework and recent works.

| Author | Year | Feature | Lead | Classifier | Detection of Onset and Ending Point | Database | No. of Class | F1-score |
|---|---|---|---|---|---|---|---|---|
| Acharya et al. (2017) | 2017 | Raw ECG | 1 | Eleven-layer Deep CNN | X | MITDB + AFDB + CUDB | 4 | 71.54 |
| Faust et al. (2018) | 2018 | RRI | 1 | LSTM | X | AFDB | 2 | 99.77 |
| Hannun et al. (2019) | 2019 | Raw ECG | 1 | ResNet | O | Clinical Database | 12 | 68.59 |
| Chen et al. (2020) | 2020 | Raw ECG + RRI | 1 | CNN + LSTM | X | MITDB | 6 | 90.82 |
| He et al. (2020) | 2020 | Raw ECG | 1 | ResNet + LSTM | X | CPSC | 9 | 80.6 |
| Ribeiro et al. (2020) | 2020 | Raw ECG | 12 | ResNet | X | S12-ECG | 6 | 92.65 |
| Teplitzky et al. (2020) | 2020 | Raw ECG | 1 | ResNet | O | Clinical Database | 14 | 70 |
| | | | | | | | 11 | 80 |
| | | | | | | | 7 | 90 |
| | | | | | | | 5 | 95 |
| Pokaprakarn et al. (2022) | 2022 | Raw ECG | 1 | CNN + LSTM | O | MITDB | 5 | 87.6 |
| Kim et al. (2022) | 2022 | Raw ECG | 1 | ResNet with SE block + biLSTM | X | MITDB | 8 | 91.69 |
| | | | | | | AFDB | 4 | 92.86 |
| Hejc et al. (2024) | 2024 | Raw EGM | 5 | D-UNet | O | IAFDB | 4 | 87.11 |
| Proposed framework | – | Raw ECG | 1 | TCN + Multiscale TIF | O | MITDB | 10 | 96.45 |
| | | | | | | AFDB | 4 | 97.57 |
| | | | | | | MADB | 8 | 86.16 |

To address these limitations, Hannun et al. [62], Teplitzky et al. [15], and Pokaprakarn et al. [63] proposed arrhythmia onset and ending point detection algorithms based on sequence-to-sequence frameworks. These studies modeled relationships between sequences within a single input and classified the last 1.0-second or 2.0-second segments. After classification, the results were aggregated to determine the occurrence times of consecutive arrhythmias. These approaches were able to identify the onset and ending points only after completing inference for the entire measurement duration. Unlike ECGs, which are recorded on the skin near the heart, Hejc et al. [57] utilized intracardiac electrograms. In contrast to the proposed framework, which can detect 10 types of arrhythmias with variable lengths, this study identified four types of regular short-interval atrial beats and their onset and ending points. They demonstrated a classification performance of 80%. However, these studies limitations, have certain limitations, including inefficient inference processes, suboptimal arrhythmia classification performance requiring bidirectional analysis, and the classification of short sections using multiple channels.

The proposed method effectively extracts and learns features from 10 classes to address these limitations. In real clinical environments, frequent inferences can lead to system overload. It is also crucial to comprehensively understand the input signal patterns. Thus, achieving high performance in arrhythmia classification by accurately interpreting input patterns while minimizing inference is essential. Although the proposed framework has increased complexity and computational load to successfully learn simultaneous patterns that existing methods failed to achieve by integrating multi-scale TIF and attention techniques, it does not show a significant difference, as it can operate within a clinically acceptable processing time compared to comparative methods. Consequently, our model outperforms state-of-the-art models in arrhythmia classification and may be better suited for real-world applications than existing models.

## 4.4  Metrics and Demetrics

The proposed framework provides several notable metrics. First, the TCN block combined with the multiscale TIF block can capture both local and global arrhythmic patterns. This capability addresses the limitations of set-level performance calculations derived from sequence-level results in previous studies, helping reduce the performance gap between sequence-level and set-level evaluations. Second, the framework mitigates the issue of continuous inference at 1-second or 1.3-second intervals, enhancing system-side efficiency. Third, the proposed framework incorporates Grad-CAM to illustrate that the model predicts areas aligning with clinically significant arrhythmia points. This feature offers clinicians interpretative insights, facilitating the integration of the model's predictions into clinical workflows.

However, the proposed framework also has some drawbacks. First, while local and global representation learning effectively captures the characteristics of arrhythmia variations over time and sudden changes, the classification performance for arrhythmias such as bigeminy and trigeminy—types that rely on historical data for accurate identification—requires improvement. Second, although the number of inferences was reduced, and there was no significant statistical difference in inference time compared to other models, the computational demand remains high due to the MHA component. Future research should explore model architectures that reduce computational overhead while maintaining performance. Third, despite the fact that the noise classification performance exhibited the highest classification performance in comparison with other models, a high level of clinical classification performance was not achieved. Consequently, there is a necessity for efforts to enhance the robustness to noise.

# 5 Conclusion

In this study, we introduced a novel CDSS for classifying multiple and multiclass arrhythmias using a proposed local-global temporal fusion network with an attention mechanism. The framework incorporates a TCN with a multiscale TIF block in both the encoder and decoder, a temporal MHA component, and a bridge component, utilizing one-lead ECG data. This design enhances the model's capability to detect and classify multiple arrhythmia types through two key strategies: integrating the TCN with multiscale TIF and merging the temporal MHA in the encoder with upsampled features in the decoder. Additionally, the proposed framework employs a combination of the multiclass weighted Dice loss function and CCE to address class imbalance issues effectively.

The results demonstrated that the proposed model outperformed UNet [47], Res-UNet [48], LinkNet [22], and TCN-LinkNet (the baseline framework for the proposed method) across all evaluation metrics. Additionally, an ablation study was conducted to assess the contribution of each architectural component. The multiscale TIF framework showed a statistically significant improvement over TCN-LinkNet and TCN-LinkNet with the bridge part. Although the temporal MHA component did not yield statistically significant improvements over TCN-LinkNet, enhancements of 3% and 5% were observed for episode and duration performances, respectively. The bridge part provided slight performance gains when combined with the multiscale TIF and temporal MHA components. Overall, the combination of the multiscale TIF block, temporal MHA component, and bridge part outperformed state-of-the-art models in multiple arrhythmia detection and classification tasks, particularly in duration and episode F1-score metrics achieved on the MITDB.

Future research will explore modifying the model structure to incorporate references to past information and enhance the efficiency of convolution filter computations. Additionally, efforts will be made to investigate lightweight algorithms, such as model compression and knowledge distillation, to further optimize the model.

We confirmed that the proposed method effectively detected and classified multiple and multiclass arrhythmias while capturing both local and global arrhythmia patterns in one-lead ECG signals. Ultimately, the proposed framework can be valuable for calculating episodes and durations in a CDSS. Cardiologists may leverage this system to plan diagnoses and treatment strategies in clinical settings. Additionally, the framework can assist clinicians by incorporating predictive information into clinical workflows and providing visual representations of the data interpreted by AI-based ECG models.

classification using convolutional recurrent neural networks," IEEE J. Biomed. Health Inform., vol. 26, no. 2, pp. 572–580, 2021.